\documentclass[letterpaper,10pt,journal,twoside]{IEEEtran}

\makeatletter
\let\NAT@parse\undefined
\makeatother

\PassOptionsToPackage{colorlinks=true,citecolor=blue,linkcolor=blue, urlcolor=black,bookmarks=false,hypertexnames=true}{hyperref}
\usepackage{hyperref}

\usepackage[table]{xcolor}
\usepackage{amsmath}
\usepackage{graphicx}
\usepackage{cleveref}
\usepackage{makecell}
\usepackage{multirow}
\usepackage{subfig}
\usepackage{pgfplots}
\usepackage{adjustbox}
\usepackage{relsize}
\usepackage{hhline}

\usepackage[utf8]{inputenc}
\usepackage{CJKutf8}

\usepackage[table]{xcolor}

\definecolor{lightgreen}{RGB}{198, 242, 126}   
\definecolor{greenish}{RGB}{17, 240, 121}    
\definecolor{lightyellow}{RGB}{252, 252, 189} 

\usepackage{tabularx,makecell,multirow,xcolor}
\newcolumntype{Y}{>{\centering\arraybackslash}X} 
\newcolumntype{Z}{>{\centering\arraybackslash}X}

\usepackage[font=footnotesize]{caption}

\usepackage{graphics}           
\usepackage{times}              
\usepackage{amsmath}            
\usepackage{amssymb}            
\usepackage{graphicx}
\usepackage{algorithm}
\usepackage[noend]{algpseudocode}
\usepackage{booktabs}
\usepackage{color}
\usepackage{multirow}
\usepackage{caption}
\usepackage{rotating}
\usepackage{cite}
\usepackage[normalem]{ulem}
\definecolor{instructioncolor}{rgb}{.5,.5,.5}

\definecolor{darkorange}{HTML}{C75A00}

\def\secref#1{Section~\ref{#1}}
\def\figref#1{Fig.~\ref{#1}}
\def\tabref#1{Table~\ref{#1}}
\def\eqref#1{(\ref{#1})}

\newcommand{\rom}[1]{\uppercase\expandafter{\romannumeral #1\relax}}

\makeatletter
\usepackage{xspace}
\DeclareRobustCommand\onedot{\futurelet\@let@token\@onedot}
\def\@onedot{\ifx\@let@token.\else.\null\fi\xspace}

\def\etal{{\textit{et al}}\onedot}
\makeatother

\usepackage{array}
\newcolumntype{L}[1]{>{\raggedright\let\newline\\\arraybackslash\hspace{0pt}}m{#1}}
\newcolumntype{C}[1]{>{\centering\let\newline\\\arraybackslash\hspace{0pt}}m{#1}}
\newcolumntype{R}[1]{>{\raggedleft\let\newline\\\arraybackslash\hspace{0pt}}m{#1}}

\newcommand{\RALReceivedDate}{November 9, 2025}
\newcommand{\RALRevisedDate}{February 8, 2026}
\newcommand{\RALAcceptedDate}{March 27, 2026}
\newcommand{\smallgap}{\vspace{-0.1cm}}

\title{GaussianFlow SLAM: Monocular Gaussian Splatting SLAM \\ Guided by GaussianFlow}
\author{
	Dong-Uk Seo$^{1}$, Jinwoo Jeon$^{1}$, \textit{Student Member, IEEE},
	Eungchang Mason Lee$^{1}$, \textit{Member, IEEE}, \\
	and Hyun Myung$^{1*}$, \textit{Senior Member, IEEE}%
	\thanks{Manuscript received: \RALReceivedDate; Revised: \RALRevisedDate; Accepted: \RALAcceptedDate.}%
	\thanks{This paper was recommended for publication by Editor Abhinav Valada upon evaluation of the Associate Editor and Reviewers' comments. This work was supported in part by the National Research Council of Science \& Technology (NST) grant by the Korea government (MSIT) (No. GTL25041-000), and in part by the Institute of Information \& Communications Technology Planning \& Evaluation (IITP) grant funded by the Korea government (MSIT) (No. RS-2025-25443318, Physically-grounded Intelligence: A Dual Competency Approach to Embodied AGI through Constructing and Reasoning in the Real World).}%
	\thanks{$^*$Corresponding author: Hyun Myung.}%
	\thanks{$^{1}$Dong-Uk Seo, Jinwoo Jeon, Eungchang Mason Lee, and Hyun Myung are with the School of Electrical Engineering, KAIST (Korea Advanced Institute of Science and Technology), Daejeon 34141, Republic of Korea (e-mail: {\tt\footnotesize\{dongukseo, jinuok, eungchang\_mason, hmyung\}@kaist.ac.kr}).}%
	\thanks{Digital Object Identifier (DOI): see top of this page.}%
}

\begin{document}
	\begin{CJK}{UTF8}{mj}
\maketitle

\begin{abstract}
  %
  Gaussian splatting has recently gained traction as a compelling map representation for SLAM systems, enabling dense and photo-realistic scene modeling. However, its application to monocular SLAM remains challenging due to the lack of reliable geometric cues from monocular input. Without geometric supervision, mapping or tracking could fall in local-minima, resulting in structural degeneracies and inaccuracies. To address this challenge, we propose GaussianFlow SLAM, a monocular 3DGS-SLAM that leverages optical flow as a geometry-aware cue to guide the optimization of both the scene structure and camera poses. By encouraging the projected motion of Gaussians, termed GaussianFlow, to align with the optical flow, our method introduces consistent structural cues to regularize both map reconstruction and pose estimation.
  Furthermore, we introduce normalized error-based densification and pruning modules to refine inactive and unstable Gaussians, thereby contributing to improved map quality and pose accuracy.
  Experiments conducted on public datasets demonstrate that our method achieves superior rendering quality and tracking accuracy compared with state-of-the-art algorithms. {The source code is available at: \url{https://github.com/url-kaist/gaussianflow-slam}.}
  

  
\end{abstract}
\begin{IEEEkeywords}
	Gaussian splatting, SLAM, optical flow
\end{IEEEkeywords}

\vspace{-0.5cm}

\section{Introduction}
\label{sec:intro}

\IEEEPARstart{V}{isual} simultaneous localization and mapping (SLAM) has long been studied as a core technology for robotics, AR/VR, and autonomous navigation. Most conventional, image-based approaches reconstruct sparse point-cloud that is sufficient for localization but not for dense geometry or photo-realistic rendering~\cite{qin2018vins, campos2021orb, lim2022uv, engel2017direct}. 
Recent differentiable scene representations, including neural radiance fields~\cite{mildenhall2021nerf, barron2022mip} and 3D Gaussian splatting (3DGS)~\cite{kerbl20233d}, enable dense, high-fidelity reconstructions. Among them, 3DGS offers real-time rasterization and orders-of-magnitude faster training than volume-based methods.
Capitalizing on the speed and differentiability gains, researchers have employed 3DGS as the primary map representation in SLAM pipelines by seeding each Gaussian with depth measurements (e.g. RGB-D, stereo, LiDAR). Such depth cues have enabled stable 3DGS-SLAM with accurate pose estimation for high-fidelity reconstructions~\cite{keetha2024splatam, yan2023gs, peng2024rtgslam, matsuki2024gaussian, lang2024gaussian}.

Despite progress in depth-supervised 3DGS-SLAM, monocular 3DGS-SLAM remains challenging due to the lack of explicit depth cues for initializing and constraining Gaussians.
As 3DGS is optimized around given 3D points, the absence of accurate geometric priors increases susceptibility to local-minima, making Gaussians fail to converge to geometrically consistent positions and instead overfit to particular views, ultimately degrading performance. 
Multi-view training could alleviate this issue~\cite{matsuki2024gaussian} but remains limited without geometric cues such as cross-view correspondences. Alternatively, depth predicted by neural networks can bootstrap and constrain Gaussians~\cite{sun2024mm3dgs, yu2025rgb, zhang2025hi, Zheng2025WildGS}, yet inaccurate priors or scale misalignment may accumulate errors or reintroduce structural distortion in the overall geometry.




\begin{figure}[t]
	\centering
	\graphicspath{{pics/intro/}} 
	\includegraphics[width=0.43\textwidth]{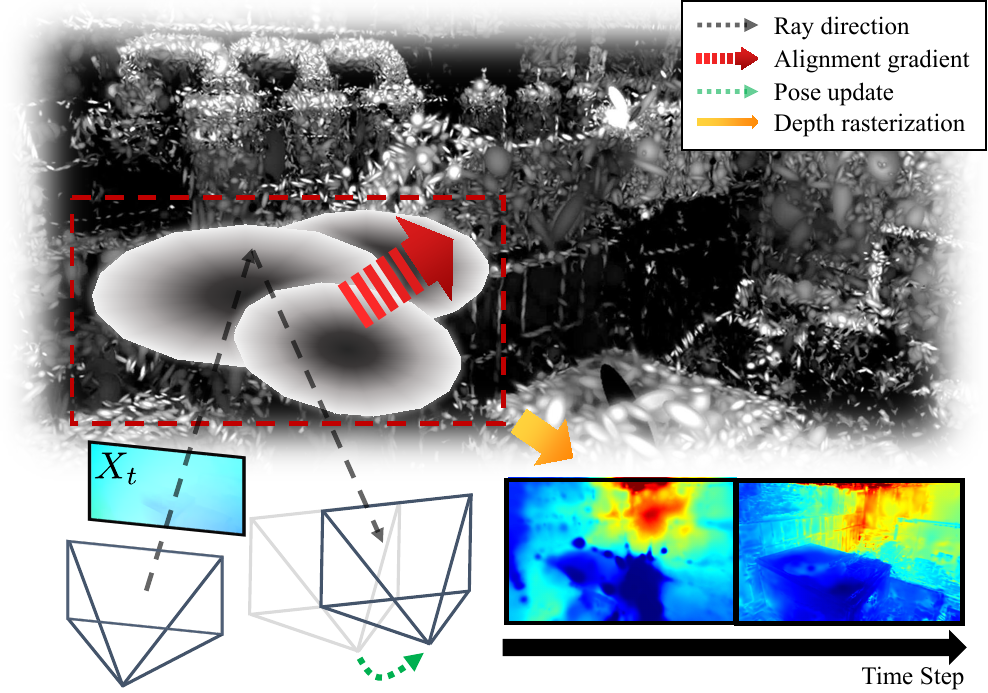}
	\caption{{Illustration of GaussianFlow SLAM process.} Our method leverages optical flow $X_t$ to optimize both 3DGS and camera pose. Along each pixel's ray direction, the projected motion of 3DGS is guided to align with the optical flow. The resulting alignment gradient enables 3DGS to iteratively update its geometric structure and the camera pose. From the rasterized depth images, it can be observed that the geometric structure of 3DGS becomes increasingly refined, recovering more valid geometric shapes over time.}
	\label{fig:intro_1}
	\vspace{-0.6cm}
\end{figure}


Another key challenge in monocular SLAM with 3DGS lies in the difficulty of leveraging the 3DGS for accurate pose estimation. Previous methods~\cite{yan2023gs, matsuki2024gaussian, schmidt2024noposegs} typically estimate the camera pose by minimizing the photometric error between the input image and a rendered image from the 3DGS.
However, when the geometry represented in the 3DGS is imprecise or locally degenerate, the rendered image becomes misaligned with the real scene, degrading pose estimation. 
This could initiate a self-reinforcing loop in which inaccurate maps degrade pose tracking, and the degraded poses in turn corrupt the map. 
To solve this problem, some methods decouple tracking with separate modules~\cite{zhang2025hi, hhuang2024photoslam, Zheng2025WildGS}, but this turns 3DGS into a passive visualizer that does not provide a feedback to improve pose estimation.


To address these challenges, we propose GaussianFlow SLAM, a monocular 3DGS-SLAM framework which leverages dense optical flow as a geometry-aware cue to guide the optimization of both the 3DGS and camera poses. Dense optical flow inherently encodes information of relative camera pose and depth, providing per-pixel correspondences. By aligning the projected motion of Gaussians, {i.e.,} GaussianFlow, with the dense optical flow, our method continuously corrects 3DGS geometry during optimization, even without direct depth measurements, as illustrated in~\figref{fig:intro_1}. 
Furthermore, our method uses GaussianFlow to guide the convergence of optical flow in pose estimation modules. This facilitates the feedback loop, allowing the mapping process to improve tracking accuracy.



Meanwhile, continuous geometry correction driven by the learning of GaussianFlow may inadvertently introduce under-densified inactive Gaussians or floaters, which are visually disturbing and structurally unstable. 
To address this problem, we propose normalized error-based densification and pruning modules that leverage per-Gaussian error to selectively adjust inactive and unstable Gaussians. These modules enhance both the quality of the map and the pose estimation accuracy in the SLAM pipeline.

In summary, our main contributions are as follows:
\begin{itemize} 
	\item To the best of our knowledge, GaussianFlow SLAM is the first 3DGS-SLAM framework to incorporate optical flow supervision with closed-form analytic gradients, implemented directly in the 3DGS kernel for optimization of geometry and camera poses. Our method drives optical flow to converge toward geometry-consistent correspondences using GaussianFlow, improving pose estimation.

	\item We propose normalized error-based densification and pruning modules that leverage per-Gaussian error to selectively adjust inactive and unstable Gaussians. This improves both map quality and pose accuracy by mitigating the accumulation of erroneous Gaussians.
	
	\item GaussianFlow SLAM achieves state-of-the-art tracking accuracy and photo-realistic reconstruction quality on the two public datasets~\cite{burri2016euroc, sturm12iros}, outperforming existing monocular Gaussian splatting SLAM baselines.
\end{itemize}
\vspace{-0.3cm}



\section{Related Work}
\label{sec:related}

\subsection{SLAM with Differentiable Rendering}\label{related_works1}


Following the success of neural radiance fields (NeRF) in achieving photo-realistic scene reconstruction, numerous works have extended NeRF~\cite{mildenhall2021nerf} to SLAM. Although more efficient encodings have been utilized in several NeRF-based SLAMs~\cite{zhu2022nice, zhu2024nicer, rosinol2023nerf}, slow rendering speed inherent to NeRF's ray-marching still remains an obstacle. Based on the fast, differentiable tile-based rasterization of 3DGS that enables high-fidelity rendering~\cite{kerbl20233d}, early 3DGS-SLAM methods primarily focused on RGB-D settings~\cite{keetha2024splatam, yan2023gs, matsuki2024gaussian, peng2024rtgslam}, where accurate depth provides a reliable basis for 3DGS initialization and optimization.
Representative pose-tracking strategies include photometric-gradient tracking~\cite{yan2023gs}, joint RGB-D optimization with on-manifold pose updates~\cite{matsuki2024gaussian}, and the iterative closest point (ICP)-based alignment~\cite{peng2024rtgslam}.




On the other hand, several works have extended 3DGS to monocular SLAM by bootstrapping Gaussians either from sparse 3D feature points~\cite{hhuang2024photoslam} or from dense depth predicted by neural networks~\cite{ zhang2025hi, yu2025rgb, Zheng2025WildGS}. However, sparse features can destabilize mapping, and depth-prior methods often suffer from scale misalignment and accumulated geometric errors.

In terms of pose tracking, MonoGS~\cite{matsuki2024gaussian} and NoPoseGS~\cite{schmidt2024noposegs} optimized photometric loss via on-manifold optimization but remained susceptible to overfitting, where errors in the 3DGS may introduce biases to subsequent pose estimates and reinforce degradation in the feedback loop. Although separate tracking modules alleviate this instability~\cite{hhuang2024photoslam, zhang2025hi, Zheng2025WildGS}, these methods reduce 3DGS to a passive renderer; in contrast, our method leverages GaussianFlow to provide geometry-aware feedback that improves tracking accuracy.

\vspace{-0.3cm}

\begin{figure*}[ht]
	\centering
	\vspace{-0.2cm}
	\graphicspath{{pics/metho/}} 
	\includegraphics[width=0.9\textwidth]{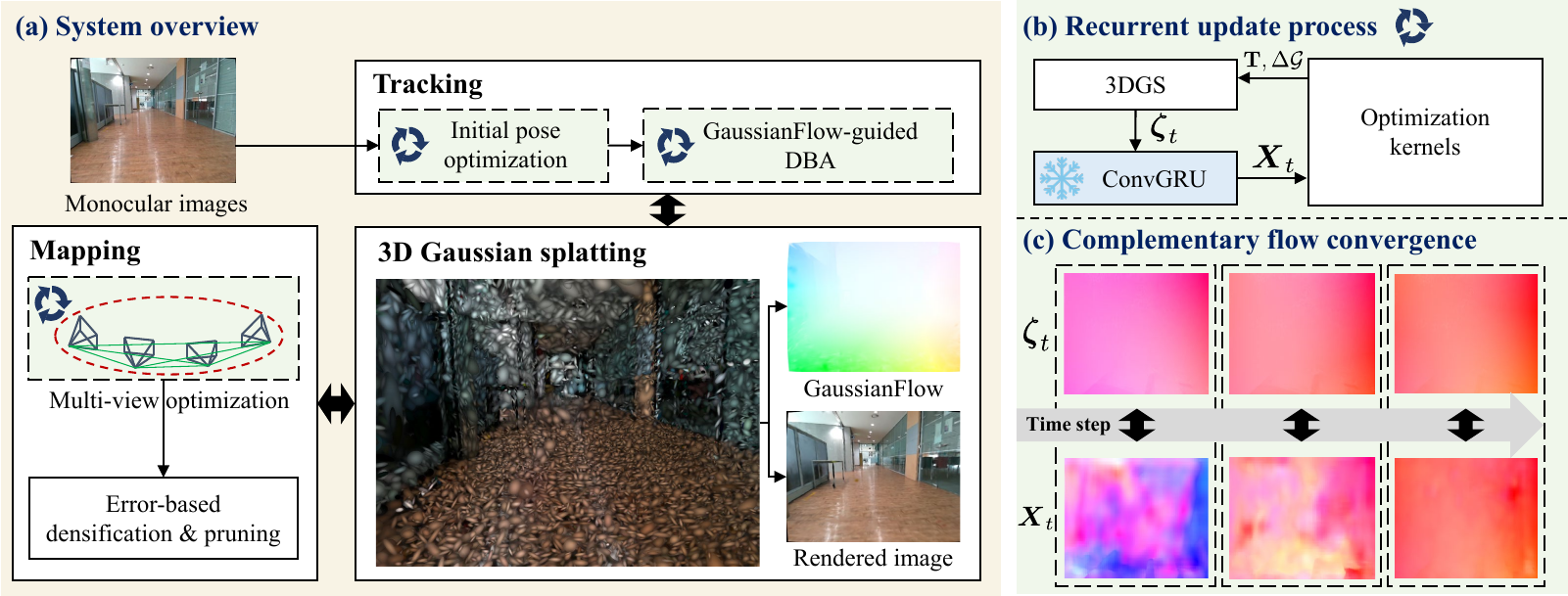}
	\caption{{(a)} The overall framework of GaussianFlow SLAM. Each of the three modules highlighted by the dashed box performs a recurrent update process during optimization.
	{(b)} The recurrent update process. The GaussianFlow $\zeta_t$ is passed to the ConvGRU (Convolutional Gated Recurrent Unit) module to predict optical flow $X_t$, which is then used by optimization kernels to update either the camera pose $\mathbf T$ or the 3DGS $\mathcal G$. The updated $\mathcal G$ are subsequently used to re-estimate $\zeta_t$. (c) Visualization of the complementary convergence between GaussianFlow $\zeta_t$ and optical flow $X_t$ over time.} 
	\label{fig:metho_overall}
	\vspace{-0.66cm}
\end{figure*}


\subsection{Optical Flow for SLAM}\label{eqn:related_works2}



Early visual SLAM systems~\cite{qin2018vins, geneva2020openvins} tracked sparse feature correspondences, often via sparse optical flow~\cite{lucas1981iterative}. 
With modern neural networks, dense optical flow has become substantially more accurate and is now widely used in SLAM, including learning-based pose estimation~\cite{wang2021tartanvo}, robust residual modeling~\cite{min2020voldor}, and recurrent refinement frameworks that achieve state-of-the-art performance~\cite{teed2021droid, teed2023deep, teed2020raft}. 
While a few studies have explored training 3DGS with optical flow, these works have focused primarily on learning dynamic content~\cite{gao2024gaussianflow, zhu2024motiongs, lin2024gaussian} or improving offline scene understanding~\cite{zhou2024hugs}. 
%
To the best of our knowledge, GaussianFlow SLAM is the first algorithm that directly incorporates optical flow supervision into 3DGS-SLAM with the closed-form analytic gradients, thereby enabling the optimization of pose and scene geometry.


\vspace{-0.4cm}



\section{GaussianFlow SLAM} 
\label{sec:metho}


\subsection{Overall Framework}\label{subsec:overall}
For incoming monocular frames, a keyframe is selected using the rough optical flow with respect to the last keyframe and it is incorporated into the optimization window, following the strategies similar to previous works~\cite{teed2021droid, zhang2025hi}.
With an initial pose estimate (\secref{section:tracking_and_mapping}), each keyframe identifies regions with insufficient Gaussian density or high rendering loss, and inserts new Gaussians into these regions. Subsequently, our system operates in an alternating loop, where dense bundle adjustment (DBA) guided by GaussianFlow and multi-view optimization using GaussianFlow loss (\secref{section:gsflow_opt}) are iteratively performed across multiple keyframes (\secref{section:tracking_and_mapping}). 
During mapping, we employ our proposed normalized error-based densification and pruning modules (\secref{metho:gs_manage}) to adjust inactive and unstable Gaussians.
An overview of the proposed system is illustrated in~\figref{fig:metho_overall}(a).

\vspace{-0.3cm}
\subsection{GaussianFlow Optimization}\label{section:gsflow_opt}
\subsubsection{GaussianFlow Loss}\label{section:closed-form}
We begin by introducing the concept of GaussianFlow, which is the projected motion of Gaussians. Using the definition proposed by Gao \textit{et al.}~\cite{gao2024gaussianflow}, GaussianFlow is expressed analogously to optical flow on the static scene, which is the motion of pixel $\mathbf p^j_{t}$ in the frame $I_t$ to the corresponding pixel $\mathbf p^j_{t+1}$ in the frame $I_{t+1}$ using 3DGS. Specifically, Gaussians rasterized at pixel $ \mathbf p^j_{t}$ are normalized to the canonical space and unnormalized to the image space of the frame $I_{t+1}$. GaussianFlow $\boldsymbol \zeta_t(\mathbf p^j_t)$ is then formulated as follows:
\vspace{-0.2cm}
\begin{equation}
\smallgap
\boldsymbol \zeta_t(\mathbf p^j_{t}) = \sum_{i \in \mathcal G^j_t} w_{ij}[\mathbf B_{i,t+1} \mathbf B_{i,t}^{-1}\boldsymbol \delta_{i,j,t} + \boldsymbol \mu_{i,t+1}] - \mathbf p^j_t,
\vspace{-0.1cm}
\end{equation}
where $\mathbf{B}_{i,t} = \boldsymbol{\Sigma}_{i,t}^{\prime 1/2}$ and $\boldsymbol{\delta}_{i,j,t} = \mathbf{p}^j_{t} - \boldsymbol{\mu}_{i,t}$; ${\mathcal G^j_t}$ denotes the set of Gaussians that contribute to the rendering of $\mathbf p^j_{t}$; the matrix $\boldsymbol \Sigma_{i,t}'$ represents the 2D covariance of Gaussian $\mathbf G_i$ projected to the frame $I_t$ by the pose $\mathbf T_t$; $\boldsymbol \mu_{i,t}$ represents the 2D center of each Gaussian $\mathbf G_i$ projected to the frame $I_t$; the weight $w_{ij}=\alpha_{ij}\prod_{k<i} (1-\alpha_{kj})$ follows the $\alpha$-blending of 3DGS~\cite{kerbl20233d}, with $\alpha_{ij}=o_i\exp(-\boldsymbol{\delta}_{i,j,t}^\top{\boldsymbol{\Sigma}^{\prime -1}_{i,t}}\boldsymbol{\delta}_{i,j,t}/2)$ and $o_i$ denotes the opacity of Gaussian $\mathbf G_i$. In the following sections, we derive equations using only consecutive frames $(I_t, I_{t+1})$ for simplicity, but the derivations can also be applied to other keyframe pairs $(I_t, I_{t+k})$.




To use GaussianFlow as the loss for optimization, prior works~\cite{gao2024gaussianflow, zhu2024motiongs, lin2024gaussian} either approximated $\mathbf B_{i,t+1} \mathbf B_{i,t}^{-1}$ as identity, or calculated the gradients for backpropagation using PyTorch's automatic differentiation~\cite{paszke2017automatic}. 
The former blocks gradient propagation to the 3DGS covariance through $\mathbf B_{i,t+1}\mathbf B_{i,t}^{-1}$, hindering GaussianFlow optimization of covariances. The latter yields correct derivatives, but it relies on a redundant dynamic computation graph and executing the backward pass in a separate PyTorch's \textit{autograd} engine, which cannot be fused into the low-level, tile-based rasterization kernel of 3DGS. As a result, {autograd} based approach limits scalability with regard to compute time and memory in SLAM settings that require repeated pose-gradient propagation and map optimization over large numbers of Gaussians.
To enable kernel-integrated and scalable optimization, we analytically decompose $\mathbf B_{i,t+1}$ and $\mathbf B_{i,t}^{-1}$ using closed-form formula of eigen decomposition~\cite{deledalle2017closed}, yielding gradients that support backpropagation through Gaussian parameters and camera poses,
which will be described in more detail in~\secref{metho:analy_gradient}.

With the GaussianFlow $\boldsymbol \zeta_t(\mathbf p^j_{t})$, we define the flow loss by comparing it against the optical flow $\boldsymbol X^j_{t}$ at pixel $\mathbf p^j_{t}$ estimated from the neural network. Considering that optical flow estimation often contains noise, we adopt the log-logistic residual model proposed by Min \textit{et al.}~\cite{min2020voldor} to robustly model the flow residual. The flow loss $\mathcal L_\text{flow}^t$ is then formulated as follows:
\smallgap
\smallgap
\begin{equation}\label{eqn:flow_loss}
\mathcal L_\text{flow}^t = -\sum_{j \in I_t} q_j(\kappa_{t}^j = 1) \log \frac{P(\boldsymbol X_{t}^j \mid \boldsymbol \zeta_t, \kappa_{t}^j = 1)}{\sum_{\kappa_{t}^j} P(\boldsymbol X_{t}^j \mid \boldsymbol \zeta_t, \kappa_{t}^j)},
\end{equation} 
\smallgap
where
\smallgap
\smallgap
\begin{equation}
P(\boldsymbol X_{t}^j | \boldsymbol \zeta_t, \kappa_t^j) = \begin{cases}
\rho(\boldsymbol \zeta_t(\mathbf p_j) ||\boldsymbol X_t^j)  & \text{if}~~\kappa_t^j = 1, \\
\nu(\boldsymbol {X}_t^j)  & \text{if}~~\kappa_t^j = 0.
\end{cases}
\smallgap
\smallgap
\end{equation}
In this formulation, $q_j(\kappa_t^j \!=\! 1)$ denotes the normalized optical flow confidence score of pixel $\mathbf p^j_t$ acquired from the optical flow network, and the binary variable $\kappa_t^j$ represents the inlier (1) or outlier (0) status of the optical flow estimate. 
 The function $\rho(\cdot|| \cdot)$ is the probability density function of the log-logistic distribution for the optical flow, while $\nu(\cdot)$ denotes the density function of a uniform distribution, used to account for outliers.

%
%
%

\subsubsection{Closed-Form Analytic Gradients}\label{metho:analy_gradient}
Our goal is to backpropagate the flow loss to both Gaussian parameters and camera poses, enabling GaussianFlow to directly refine geometry and pose within the 3DGS optimization.
The key challenge is that GaussianFlow depends on the projected covariance $\boldsymbol\Sigma'_{i,t}$ through the matrix functions $\boldsymbol\Sigma'^{\pm 1/2}$ (via $\mathbf M_{i,t} \!\!=\!\! \mathbf B_{i,t+1}\mathbf B^{-1}_{i,t}$), whose derivatives are not straightforward to implement in the fused GPU kernel.
Because $\boldsymbol\Sigma'_{i,t}$ is a $2{\times}2$ symmetric positive definite matrix, we express these matrix functions via the eigen decomposition $\boldsymbol\Sigma'_{i,t}\!=\!\mathbf Q_{i,t}\mathbf S_{i,t}\mathbf Q_{i,t}^{\top}$, which yields closed-form derivatives w.r.t.\ $\mathbf Q_{i,t}$ and $\mathbf S_{i,t}$, as given in~\eqref{eqn:eigen_gradient1} and~\eqref{eqn:eigen_gradient2}. Moreover, the closed-form eigen decomposition provides analytic Jacobians from $(\mathbf Q_{i,t},\mathbf S_{i,t})$ back to $\boldsymbol\Sigma'_{i,t}$, so $\partial \mathcal L/\partial \boldsymbol\Sigma'_{i,t}$ can be obtained via the chain rule in the kernel.
With these closed-form derivatives, the flow loss gradient can be propagated to camera poses directly within the 3DGS GPU kernel, as given in~\eqref{eqn:pose_gradient1} and~\eqref{eqn:pose_gradient2}.



%

From~\eqref{eqn:flow_loss}, the per-pixel flow loss $\boldsymbol \psi^j_t$ without confidence $q_j$ for pixel $\mathbf p^j_t$ is represented as follows:
\smallgap
\smallgap
\begin{equation}\label{eqn:flow_loss_pixel}
\boldsymbol \psi^j_t = -\log \frac{P(\boldsymbol X_{t}^j \mid \boldsymbol \zeta_t, \kappa_{t}^j \!=\! 1)}{\sum_{\kappa_{t}^j} P(\boldsymbol X_{t}^j \mid \boldsymbol \zeta_t, \kappa_{t}^j)}.
\end{equation}
Because~\eqref{eqn:flow_loss_pixel} is differentiable and parallelizable, gradients of $\mathcal L^t_\text{flow}$ with respect to the 3DGS parameters could be derived via the chain rule in the kernel. Specifically, we derive the gradients of the flow loss with respect to the 3D mean $\mathbf x_{i,t}$ and the covariance $\boldsymbol \Sigma_{i,t}$ as follows:

\smallgap
\smallgap
{\setlength{\abovedisplayskip}{4pt}%
	\setlength{\belowdisplayskip}{4pt}%
	\small
\begin{equation}\label{eqn:gradient_mu}
\frac{\partial \mathcal L_\text{flow}^t}{\partial \mathbf x_{i,t}} 
= \frac{\partial \mathcal L_\text{flow}^t}{\partial \boldsymbol \mu_{i,t}}
\frac{\partial \boldsymbol \mu_{i,t}}{\partial \mathbf x_{i,t}}
+ \frac{\partial \mathcal L_\text{flow}^t}{\partial \boldsymbol \mu_{i,t+1}}
\frac{\partial \boldsymbol \mu_{i,t+1}}{\partial \mathbf x_{i,t}}
+ \frac{\partial \mathcal L_\text{flow}^t}{\partial \boldsymbol{\Sigma}_{i,t}}
\frac{\partial \boldsymbol{\Sigma}_{i,t}}{\partial \mathbf x_{i,t}},
\smallgap
\end{equation}
\smallgap
\begin{equation}\label{eqn:gradient_sigma}
\frac{\partial \mathcal L_{\text{flow}}^t}{\partial \boldsymbol{\Sigma}_{i,t}}
= \sum_{\ell \in \{t, t+1\}} \left(
\frac{\partial \mathcal L_{\text{flow}}^t}{\partial \mathbf Q_{i,\ell}}
\frac{\partial \mathbf Q_{i,\ell}}{\partial \boldsymbol{\Sigma}_{i,t}}
+ \frac{\partial \mathcal L_{\text{flow}}^t}{\partial \mathbf S_{i,\ell}^{s(\ell)}}
\frac{\partial \mathbf S_{i,\ell}^{s(\ell)}}{\partial \boldsymbol{\Sigma}_{i,t}}
\right),
\end{equation}
}where $s(t)\!\!=\!\!-1/2$ and $s(t+1)\!\!=\!\!1/2$. Note that $\mathcal L^t_{\text{flow}}$ also differentiates through $w_{ij}$; the resulting gradient terms follow the standard 3DGS backward pass, and we omit their derivation for brevity.
The gradient with respect to the opacity $o_i$ is obtained directly via the scalar chain rule. 
Subsequently, 
we derive the following gradients, when the GaussianFlow loss gradient is denoted as $\mathbf D_t = {\partial \mathcal L_\text{flow}^t} / {\partial \boldsymbol \zeta_t}$:

{\setlength{\abovedisplayskip}{4pt}%
	\setlength{\belowdisplayskip}{4pt}%
	\small
	\smallgap
	\smallgap
	\smallgap
	\smallgap
	\begin{align}
	\frac{\partial \mathcal L_\text{flow}^t}{\partial \boldsymbol \mu_{i,\ell}} 
	&= 
	\begin{cases}
	\langle -\mathbf M_{i,t}^\top \mathbf D_t \rangle_{\Omega_i}, & \ell = t, \\
	\langle \mathbf D_t \rangle_{\Omega_i}, & \ell = t+1 ,
	\end{cases} \\
	\frac{\partial \mathcal L_\text{flow}^t}{\partial (\mathbf B_{i,t}^{-1}, \mathbf B_{i,t+1})}
	&= \Big\langle \big( \mathbf B_{i,t+1}\mathbf v_{i,t},\; \mathbf v_{i,t}\mathbf B_{i,t}^{-\top} \big) \Big\rangle_{\Omega_i}.
	\end{align}
}where $\mathbf v_{i,t} = \mathbf D_t(\mathbf p_t^j - \boldsymbol \mu_{i,t})^\top,$ and $\langle f \rangle_{\Omega_i} := \sum_{j \in \Omega_i} w_{ij} f(j)$ denotes the weighted summation over the pixel set $\Omega_i$ rasterized by $\mathbf G_i$. Finally, letting $\mathbf A_t \!\!=\!{\partial \mathcal L_\text{flow}^t} / {\partial \mathbf B_{i,t}^{-1}}$ and $\mathbf A_{t+1} \!\!=\!{\partial \mathcal L_\text{flow}^t} / {\partial \mathbf B_{i,t+1}}$, the gradients with respect to the eigen decomposition are derived as follows:



{\setlength{\abovedisplayskip}{4pt}%
	\setlength{\belowdisplayskip}{4pt}%
	\small
	\smallgap
	\smallgap
\begin{align}
\frac{\partial \mathcal L_\text{flow}^t}{\partial \mathbf Q_{i,\ell}} &= (\mathbf A_\ell + \mathbf A_\ell^\top)\mathbf Q_{i,\ell}\mathbf S_{i,\ell}^{s(\ell)}, \label{eqn:eigen_gradient1} \\
\frac{\partial \mathcal L_\text{flow}^t}{\partial \mathbf S_{i,\ell}^{s(\ell)}} &= \mathbf Q_{i,\ell}^\top \mathbf A_\ell \mathbf Q_{i,\ell}. \label{eqn:eigen_gradient2}
\end{align}
}These results are then propagated to the individual elements of $\boldsymbol{\Sigma}_{i,t}$ through the scalar chain rule, enabling optimization of the covariance.

Meanwhile, by parameterizing the camera poses $\mathbf T \in \textit{SE}(3)$ using the Lie algebra $\mathfrak{se}(3)$\cite{sola2018micro}, we extend our derivation to include gradients with respect to the camera poses. For notational simplicity, the gradient for the pose is expressed as ${\partial \mathcal L} / {\partial \mathbf T}$.
Using the gradient from GaussianFlow loss, we derive the gradients for each poses as follows:

\vspace{-0.4cm}
{
	\setlength{\belowdisplayskip}{4pt}%
	\small
\begin{align}\label{eqn:pose_gradient1}
\frac{\partial \mathcal L_\text{flow}^t}{\partial \mathbf T_t} &= \sum_i \left(\frac{\partial \mathcal L_{\text{flow}}^t}{\partial  \mathbf x_{i,t}} \frac{\partial \mathbf x_{i,t}}{\partial \boldsymbol \mu_{i,t}} \frac{\partial \boldsymbol \mu_{i,t}}{\partial \mathbf T_t} + \frac{\partial \mathcal L_{\text{flow}}^t}{\partial \boldsymbol \Sigma_{i,t}'} \frac{\partial \boldsymbol \Sigma_{i,t}'}{\partial \mathbf T_t} \right), \\ \label{eqn:pose_gradient2}
\frac{\partial \mathcal L_\text{flow}^t}{\partial \mathbf T_{t+1}} &= \sum_i \left( \frac{\partial \mathcal L_{\text{flow}}^t}{\partial \mathbf x_{i,t}} \frac{\partial \mathbf x_{i,t}}{\partial \boldsymbol \mu_{i,t+1}} \frac{\partial \boldsymbol \mu_{i,t+1}}{\partial \mathbf T_{t+1}} + \frac{\partial \mathcal L_{\text{flow}}^t}{\partial \boldsymbol \Sigma_{i,t+1}'} \frac{\partial \boldsymbol \Sigma_{i,t+1}'}{\partial \mathbf T_{t+1}} \right).
\end{align}
}These formulations enable pose updates not only for the frame $I_t$ where rasterization occurs, but also for the target frame $I_{t+1}$ of the optical flow. 
\vspace{-0.4cm}
\subsection{Tracking and Mapping}\label{section:tracking_and_mapping}


As described in~\secref{subsec:overall}, our tracking strategy employs two modules. To estimate the initial pose of a new keyframe $I_{t+1}$, we use the GaussianFlow loss \eqref{eqn:flow_loss} between $I_{t+1}$ and the two preceding keyframes $I_{t-1}$ and $I_t$, along with the photometric loss $\mathcal{L}_\text{image}^{t+1}$ for $I_{t+1}$, which exploits the image loss gradient ${\partial \mathcal L_\text{image}} / {\partial \mathbf T}$ derived in~\cite{matsuki2024gaussian, schmidt2024noposegs}. 
Although geometric optimization has not yet been performed for $I_{t+1}$, the preceding keyframes have already been optimized, enabling the usage of GaussianFlow loss to update the pose of $I_{t+1}$. Accordingly, the loss function for initial pose tracking is formulated as follows:
\smallgap
{
	\setlength{\belowdisplayskip}{4pt}%
\begin{align}\label{eqn:init_pose}
\min_{\substack{\mathbf{T}_{t+1} \in \textit{SE}}(3)} \mathcal{L}_\text{image}^{t+1} + \lambda_1 (\mathcal{L}_\text{flow}^t + \mathcal{L}_\text{flow}^{t-1}),
\end{align}
}where $\lambda_1$ is flow loss weight and $\mathcal L_\text{image}$ denotes the original 3DGS loss, a weighted sum of L1 loss and structural dissimilarity (DSSIM) loss~\cite{kerbl20233d}.

After the initial pose estimation, we optimize the poses of multiple keyframes in the window via DBA~\cite{teed2021droid} guided by GaussianFlow. 
Because first-order optimization in 3DGS lacks curvature information, it can struggle to converge in highly non-linear multi-pose refinement~\cite{triggs1999bundle}. We therefore adopt the DBA for second-order optimization, which solves a flow-based bundle adjustment problem between keyframes. To leverage cross-view consistency, we use GaussianFlow as the flow input: $\zeta_t$ is fed into the pretrained ConvGRU~\cite{teed2021droid} to produce $X_t$, which in turn guides the optimization.



During mapping, the poses estimated by DBA are held fixed while we perform multi-view optimization of the 3DGS $\mathcal G$. Within the optimization window $\mathcal W$, optical flow-based edges are connected between keyframes, following a similar scheme to~\cite{teed2021droid, zhang2025hi}.
For each mapping iteration, one of the edges connected to the keyframe $k \in \mathcal W$ is selected to compute the mapping loss, which is formulated as follows:

\smallgap
\smallgap
\smallgap
\smallgap
\smallgap
\vspace{-0.03cm}
{
	\setlength{\belowdisplayskip}{4pt}%
\begin{align}\label{eqn:mapping}
\min_{\substack{\mathcal G}}
\mathcal L_\text{image}^{t_k} + \lambda_{2} \mathcal L_\text{flow}^{t_k} + \lambda_{3} \mathcal L_\text{iso} + \lambda_4 \mathcal L_\text{opa},
\end{align} 
}where $\lambda_2, \lambda_3,$ and $\lambda_4$ are weighting coefficients for the flow, isotropic, and opacity losses, respectively. 
The isotropic loss $\mathcal{L}_\text{iso}$~\cite{matsuki2024gaussian}
and opacity entropy loss $\mathcal L_\text{opa}$~\cite{duan20244d} regularize the Gaussians to avoid excessive elongation and vague floaters. Subsequently, the system alternates between DBA pose estimation and mapping.

For these three modules (initial pose estimation, GaussianFlow-guided DBA, multi-view mapping), a recurrent update process is employed. Prior works such as RAFT~\cite{teed2020raft} and DROID-SLAM~\cite{teed2021droid} have shown that optical flow becomes more accurate through iterative refinement, and that incorporating per-keyframe projected depth further improves the accuracy of both depth and pose estimation.
Beyond the per-keyframe depth optimization, we refine optical flow using GaussianFlow rendered from a shared and continuously optimized 3DGS map. 
As illustrated in~\figref{fig:metho_overall}(b), the recurrent update process jointly refines GaussianFlow and optical flow from ConvGRU under the updated poses and the shared map, as shown in~\figref{fig:metho_overall}(c).

\vspace{-0.4cm}
\subsection{Gaussian Management and Filtering Modules}\label{metho:gs_manage}

\subsubsection{Error-based Densification and Pruning} 

\begin{figure}[t]
	\centering
	\graphicspath{{pics/exp/}} 
	\includegraphics[width=0.43\textwidth]{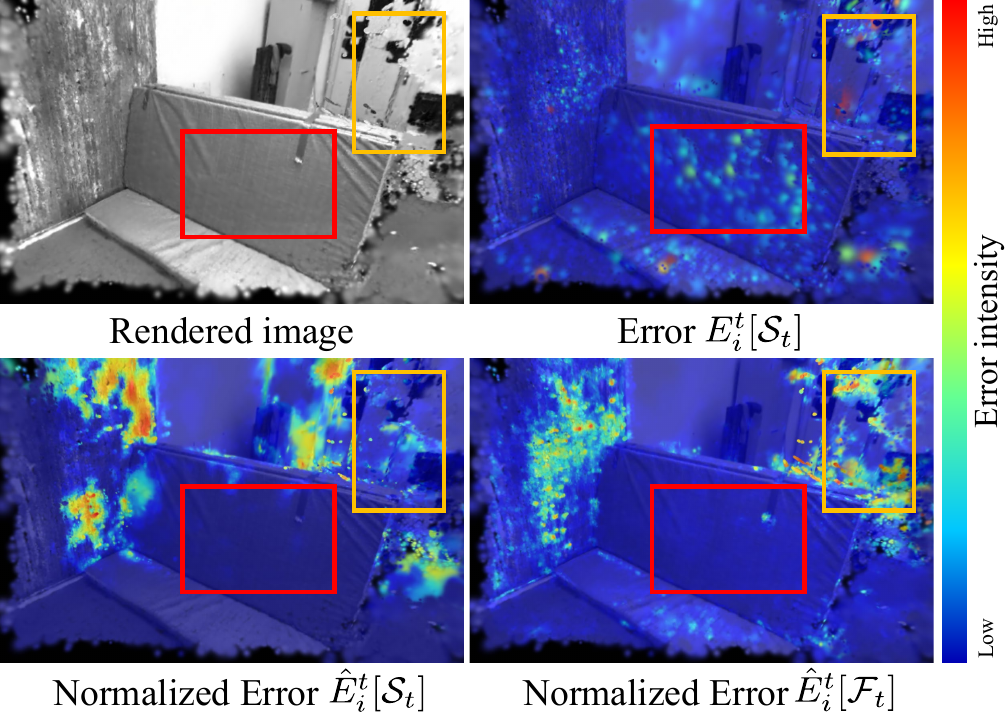} 
	\caption{{Visualization of per-Gaussian error types for the rendered image from an alternative viewpoint.} For $E_i^t[\mathcal S_t]$, the density contribution of each Gaussian is not considered, so the errors tend to highlight spatially extended Gaussians. In contrast, $\hat E_i^t[\mathcal S_t]$ and $\hat E_i^t[\mathcal F_t]$ account for the density contribution of each Gaussian, enabling delicate per-Gaussian error analysis. 
	}
\label{fig:method_error}
\vspace{-0.7cm}
\end{figure}

While new Gaussians are inserted for each keyframe, the 3DGS map can still be too sparse to capture fine textures. The original 3DGS~\cite{kerbl20233d} therefore densifies Gaussians when their position gradients exceed a threshold, but this approach may miss Gaussians that, as observed in~\cite{rota2024revising}, remain nearly stationary yet incur large errors due to local minima.
To address this limitation, we reimplemented the per-Gaussian error proposed in~\cite{rota2024revising} and propose a \emph{normalized} error for more selective densification. For a Gaussian $\mathbf G_i$ rasterized in the frame $I_t$, the Gaussian error $E_i^t$~\cite{rota2024revising} is computed as follows:
 \vspace{-0.1cm}
 {
 	\setlength{\abovedisplayskip}{6pt}%
 	\setlength{\belowdisplayskip}{3pt}%
\begin{align}\label{eqn:per_gaussian_error}
E_i^t[\mathcal E_t] = \sum_{j \in I_t} w_{ij} \mathcal{E}_t(\mathbf{p}^j),
\end{align} }where $\mathcal{E}_t$ denotes a rendering-based loss map function. 
Unlike the prior error-based approach~\cite{rota2024revising} that uses per-Gaussian error $E_i^t$, we normalize it to remove a screen-space coverage bias: $E_i^t$ grows with the number of affected pixels and thus takes larger values for wide Gaussians. We therefore divide $E_i^t$ by the Gaussian’s density contribution, yielding a scale-invariant score across different radii.
Specifically, we apply the silhouette image $\mathcal H_t$~\cite{keetha2024splatam}, which represents the per-pixel scene density, to $\mathcal{E}_t$ in~\eqref{eqn:per_gaussian_error} to compute the density contribution of each Gaussian as $D_i^t \!=\! E_i^t[\mathcal H_t]$. Then, we define the normalized error as $\hat{E}_i^t[\mathcal E_t] \!\!=\!\! {E_i^t}[\mathcal E_t]/{D_i^t}$.
For these purposes, we employ two types of loss maps, with $\mathcal S_t$ denoting the per-pixel DSSIM loss map and $\mathcal F_t(\mathbf p_t^j) \!=\! q_j(\kappa_t^j\!=\!1)\psi_t^j$ denoting the per-pixel flow loss map. 
As shown in~\figref{fig:method_error}, $E_i^t$ tends to to be larger for wide-coverage Gaussians, whereas $\hat{E}_i^t$ highlights errors more consistently with their relevance for each loss.




%
%

Leveraging both the Gaussian error $E_i^t$ and the normalized Gaussian error $\hat{E}_i^t$, 
we perform densification by splitting Gaussians within the following masks:

\vspace{-0.45cm}
{
	\setlength{\belowdisplayskip}{3pt}%
	\small
\begin{align}
M^t_1 &= \{\mathbf G_i | E_i^t[\mathcal S_t] > \eta_{s_1} \cap r_i^t > \eta_{r_1} \cap \nabla_p \mathcal G_i^t < \eta_{g_1} \}, \\
M^t_2 &= \{\mathbf G_i | \hat E_i^t[\mathcal S_t] > \eta_{s_2} \cap  r_i^t > \eta_{r_1} \}, \\
M^t_3 &= \{\mathbf G_i | r_i^t > \eta_{r_2} \}, 
\end{align}
}where $r_i^t$ is the maximum projected radius of $\mathbf G_i$ on $I_t$, $\nabla_p$ denotes the position gradient, and $\eta_{(\cdot)}$ are thresholds. 
Here, $M_1^t$ targets large Gaussians with high errors but small position gradients, $M_2^t$ targets large Gaussians with high normalized errors, and $M_3^t$ splits overly large Gaussians.
Together, $M_1^t$ and $M_2^t$ enable densification for both nearly stationary Gaussians with large errors and Gaussians with high normalized errors, mitigating the coverage bias of summed per-Gaussian error.



Meanwhile, the GaussianFlow-based mapping may shift and deform Gaussians for geometry correction, occasionally shrinking them so that they receive negligible gradient updates and become \textit{floaters} that stagnate and degrade geometric fidelity.
To remove these unstable Gaussians, we prune those identified by the following masks:

\vspace{-0.45cm}
{
	\setlength{\belowdisplayskip}{3pt}%
	\small
\begin{align}
M^t_4 &= \{\mathbf G_i | (\hat E_i^t[\mathcal S_t] > \eta_{p_1} \cup \hat E_i^t[\mathcal F_t] > \eta_{p_2})\cap r_i^t < \eta_{r_3} \}, \\ 
M^t_5 &= \{\mathbf G_i | \hat E_i^t[\mathcal S_t] > \eta_{p_3} \cap r_i^t < \eta_{r_4} \}, \\
M^t_6 &= \{\mathbf G_i | o_i^t < \eta_{o_1} \},
\end{align}	
}where $o_i^t$ is the opacity of $\mathbf G_i$ used to rasterize the frame $I_t$.
Here, $M^t_4$ removes small Gaussians with high normalized DSSIM or flow errors, $M^t_5$ extends this criterion to a wider radius range using DSSIM, and $M^t_6$ prunes low-opacity Gaussians.
Beyond pruning by size and opacity, our method performs more fine-grained pruning by removing small-radius Gaussians with high normalized DSSIM or flow errors, which frequently emerge as floaters during GaussianFlow optimization.



\subsubsection{Gaussian ID Assignment for Adaptive Window}
To maintain map consistency with a shared global 3DGS map, we optimize recent keyframes together with earlier covisible keyframes. Similar to loop-aware keyframe management in the prior approach~\cite{kong2024dgs} using Gaussian IDs, we store a single keyframe ID per Gaussian and assign it at first rasterization (insertion or split) using the ID of the rasterizing keyframe. Although approximate, the large and diverse set of Gaussians captures most covisible keyframes, providing a functionality similar to loop-closure with 3DGS that improves large-scale accuracy.
\vspace{-0.5cm}

\section{Experimental Results}
\label{sec:exp}


\subsection{Dataset}

To evaluate GaussianFlow SLAM, we conducted experiments on two public real-world datasets, the TUM RGB-D dataset~\cite{sturm12iros} and the EuRoC dataset~\cite{burri2016euroc}. TUM RGB-D provides small-scale indoor sequences that have been widely adopted for 3DGS-SLAM evaluation, where most existing monocular 3DGS-SLAM methods already have demonstrated relatively strong performance. In contrast, EuRoC dataset consists of large-scale UAV sequences with challenging motions and lighting conditions, where 3DGS-SLAM systems are more prone to geometric errors and local-minima issues, thus providing a challenging benchmark. Both benchmarks are predominantly static. Evaluating and extending our method to highly dynamic scenes is left for future work.


\vspace{-0.4cm}
\subsection{Implementation Details}\label{imple_details} 
Our method operates on monocular camera input. SLAM modules were implemented in Python, but GaussianFlow rasterization process, flow loss computation, and gradient backpropagation were implemented in the CUDA kernel. All experiments were conducted with an AMD Ryzen Threadripper 3.6 GHz PRO 5975WX processor and a single NVIDIA RTX A6000 GPU. We empirically set the hyperparameters as follows. For the GaussianFlow loss, we set $\lambda_1\!\!=\!1.0$, $\lambda_2\!\!=\!0.5$ for EuRoC and $0.1$ for TUM RGB-D, $\lambda_3\!\!=\!1.0$, and $\lambda_4\!\!=\!0.01$. The error-based densification module used $\eta_{s_1}\!\!=\!0.2$, $\eta_{r_1}\!\!=\!10$, $\eta_{g_1}\!\!=\!0.0001$, $\eta_{s_2}\!\!=\!0.1$, and $\eta_{r_2}\!\!=\!40$. For the pruning module, we set $\eta_{p_1}\!\!=\!0.6$, $\eta_{p_2}\!\!=\!0.2$, $\eta_{r_3}\!\!=\!5$, $\eta_{p_3}\!\!=\!1.5$, and $\eta_{r_4}\!\!=\!7$, and additionally used $\eta_{o_1}\!\!=\!0.05$. $\lambda_2$ was set to the larger value for EuRoC to assign a stronger weight to the flow loss, which facilitates spatial expansion in large-scale scenes.

\pgfplotsset{compat=1.13}
\begin{figure}[t]
	\centering
	\graphicspath{{pics/exp/}} 
	\includegraphics[width=0.43\textwidth]{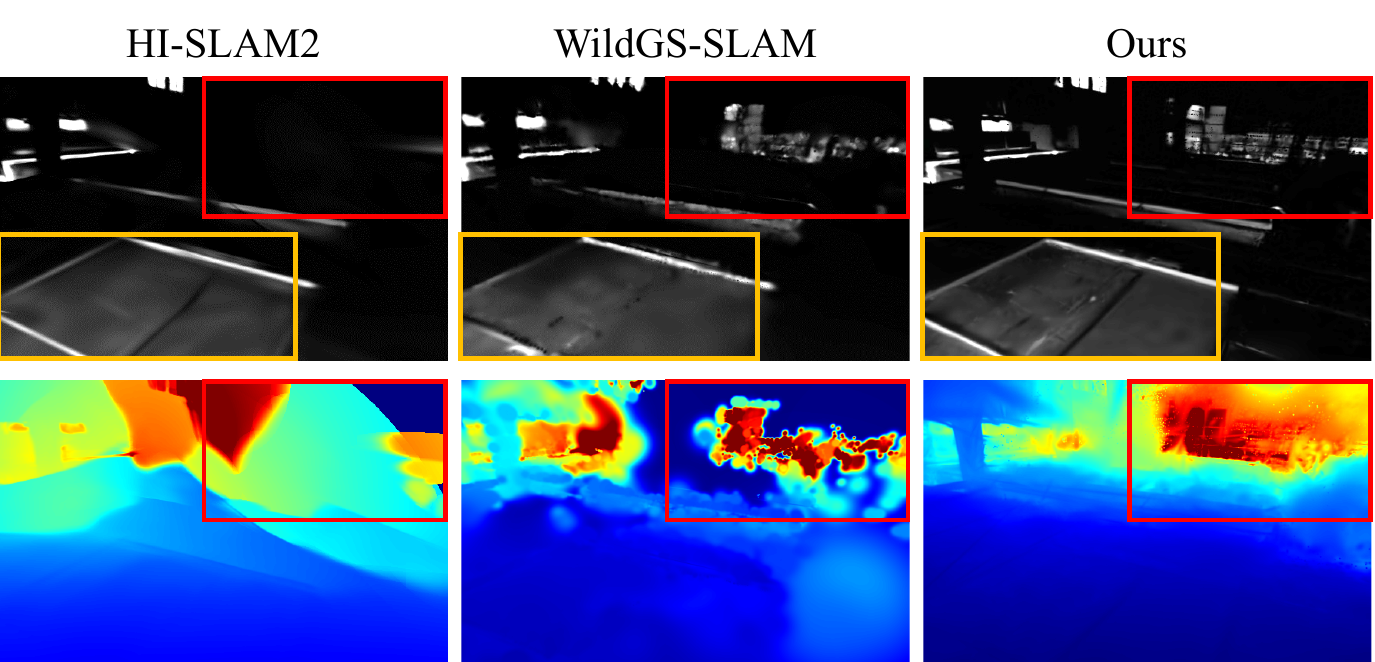}
	\caption{{Comparison of rendered images and rasterized depths for a challenging scene.} Methods exploiting monocular depth priors, such as HI-SLAM2~\cite{zhang2025hi} and WildGS-SLAM~\cite{Zheng2025WildGS}, learn incorrect 3DGS geometry when the depth prior is unreliable. In contrast, our GaussianFlow-based method robustly learns geometrically valid structures.} 
	\label{fig:result2}
	\vspace{-0.65cm}
\end{figure}

\vspace{-0.4cm}

\subsection{Baseline Algorithms}
We compare our method against publicly available 3DGS-SLAMs with monocular mode, including MonoGS~\cite{matsuki2024gaussian}, MM3DGS-SLAM~\cite{sun2024mm3dgs}, Photo-SLAM~\cite{hhuang2024photoslam}, HI-SLAM2~\cite{zhang2025hi} and WildGS-SLAM~\cite{Zheng2025WildGS}. 
Considering that our algorithm inherently provides functionality similar to loop closure, we conducted all experiments with the loop closure enabled when it is supported. 
For a fair comparison, offline full pose-graph optimization was disabled, and the map refinement was performed the same number of times across all methods.
Because MM3DGS-SLAM~\cite{sun2024mm3dgs} requires a depth image of the first frame for scale alignment, we replaced it with a monocular prior. WildGS-SLAM~\cite{Zheng2025WildGS} was run in static mode due to degraded performance in dynamic mode. 
\vspace{-0.4cm}



\subsection{Quantitative Comparison}
\tabref{tab:rmse_main} reports pose accuracy measured by RMSE of absolute trajectory error (ATE). While most methods perform reliably on room-scale TUM RGB-D, several degrade on the larger-scale EuRoC: MonoGS is prone to local minima, MM3DGS-SLAM suffers from scale misalignment, and WildGS-SLAM is affected by depth-prior errors. HI-SLAM2 alleviates these issues via robust scale alignment,  yet it still fell short compared with Photo-SLAM which is based on ORB-SLAM3~\cite{campos2021orb}, and our approach. Photo-SLAM outperforms ours on a few challenging sequences (EuRoC MH04/MH05 and TUM fr1/desk), where extremely low illumination or transient motion blur can make dense correspondences from optical flow unreliable. Nevertheless, our method achieves the best results on most sequences, and dense pixel supervision is particularly well-suited for improving rendering quality of 3DGS by providing more geometric information than sparse feature measurements.

\begin{table}[t]
	\centering
	\caption{{Tracking performance comparison on EuRoC and TUM RGB-D datasets (RMSE ATE [m]).}\setlength{\fboxsep}{1pt} The best three results are highlighted in order of the \colorbox{greenish!40}{\textbf{first}}, \colorbox{lightgreen!40}{second}, and \colorbox{lightyellow}{third}, respectively. When pose estimation diverged during the process, the result was indicated by $\times$.}
	\label{tab:rmse_main}
	\scriptsize
	\renewcommand{\arraystretch}{1.1}
	\setlength\tabcolsep{1.5pt}
	\resizebox{\columnwidth}{!}{%
			\begin{tabularx}{\columnwidth}{c|c|*{6}{>{\centering\arraybackslash}X}}
			\Xhline{4\arrayrulewidth}
			\makecell[c]{\textbf{Dataset}} & \makecell[c]{\textbf{Seq.}}
&\tiny\larger[1] \makecell[c]{MonoGS\\\cite{matsuki2024gaussian}}
&\tiny\larger[1] \makecell[c]{MM3DGS-\\SLAM~\cite{sun2024mm3dgs}}
&\tiny\larger[1] \makecell[c]{HI\text{-}SLAM2\\\cite{zhang2025hi}}
&\tiny\larger[1] \makecell[c]{WildGS-\\SLAM~\cite{Zheng2025WildGS}}
&\tiny\larger[1] \makecell[c]{Photo-\\SLAM~\cite{hhuang2024photoslam}}
&\tiny\larger[1] \makecell[c]{\textbf{Ours}} \\
			\Xhline{2\arrayrulewidth}
			\multirow{11}{*}{\makecell{EuRoC\\~\cite{burri2016euroc}}}
			& MH01 & 3.202 & 3.110 & \cellcolor{lightyellow}0.046 & 0.279 & \cellcolor{lightgreen!40}0.043 & \cellcolor{greenish!40}\textbf{0.027} \\
			& MH02 & 2.877 & 2.996 & \cellcolor{lightgreen!40}{0.036} & 0.368 & \cellcolor{lightgreen!40}{0.036} & \cellcolor{greenish!40}\textbf{0.035} \\
			& MH03 & 2.452 & $\times$ & \cellcolor{lightyellow}0.065 & 0.858 & \cellcolor{lightgreen!40}0.038 & \cellcolor{greenish!40}\textbf{0.033} \\
			& MH04 & 4.000 & $\times$ & \cellcolor{lightyellow}0.080 & 2.998 & \cellcolor{greenish!40}\textbf{0.057} & \cellcolor{lightgreen!40}0.067 \\
			& MH05 & 3.417 & 4.939 & \cellcolor{lightyellow}0.058 & 3.452 & \cellcolor{greenish!40}\textbf{0.048} & \cellcolor{lightgreen!40}0.053 \\
			& V101 & 1.650 & $\times$ & \cellcolor{lightgreen!40}0.088 & 0.149 & \cellcolor{lightyellow}0.096 & \cellcolor{greenish!40}\textbf{0.079} \\
			& V102 & 1.895 & 1.771 & \cellcolor{lightyellow}0.069 & 0.163 & \cellcolor{lightgreen!40}0.061 & \cellcolor{greenish!40}\textbf{0.050} \\
			& V103 & 1.326 & $\times$ & \cellcolor{lightyellow}0.076 & 0.160 & \cellcolor{lightgreen!40}0.065 & \cellcolor{greenish!40}\textbf{0.051} \\
			& V201 & 2.152 & $\times$ & \cellcolor{lightgreen!40}0.071 & \cellcolor{lightyellow}0.133 & $\times$ & \cellcolor{greenish!40}\textbf{0.050} \\
			& V202 & 1.195 & 1.897 & \cellcolor{lightyellow}0.065 & 0.155 & \cellcolor{lightgreen!40}0.057 & \cellcolor{greenish!40}\textbf{0.040} \\
			& V203 & 1.017 & $\times$ & \cellcolor{lightgreen!40}{0.069} & \cellcolor{lightyellow}0.214 & $\times$ & \cellcolor{greenish!40}\textbf{0.066} \\
			\Xhline{2\arrayrulewidth}
			\multirow{3}{*}{\makecell{TUM\\RGB-D\\~\cite{sturm12iros}}}
			& f1/desk   & 0.026 & 0.439 & \cellcolor{lightgreen!40}0.021 & 0.026 & \cellcolor{greenish!40}\textbf{0.015} & \cellcolor{lightyellow}0.024 \\
			& f2/xyz    & 0.047 & 0.236 & \cellcolor{lightgreen!40}{0.002} & 0.023 & \cellcolor{lightyellow}0.003 & \cellcolor{greenish!40}\textbf{0.001} \\
			& f3/office & 0.027 & 1.481 & \cellcolor{lightgreen!40}{0.017} & 0.130 & \cellcolor{lightyellow}0.021 & \cellcolor{greenish!40}\textbf{0.015} \\
			\Xhline{4\arrayrulewidth}
		\end{tabularx}
	} 
	\vspace{-0.6cm}
\end{table}

\tabref{tab:quantitative_render} reports the rendering quality evaluation results, evaluated on one out of every five non-keyframes using peak signal-to-noise ratio (PSNR), structural similarity index (SSIM), and learned perceptual image patch similarity (LPIPS), following the protocol in~\cite{matsuki2024gaussian, zhang2025hi, Zheng2025WildGS}. Our method achieved higher performance than most of the compared algorithms. 
HI-SLAM2 achieved higher PSNR and SSIM by suppressing noise through smoothing loss, whereas our method without smoothing was lower on some sequences.
However, in LPIPS, which reflects perceptual quality, our approach outperformed in most cases.

\begin{table*}[t]
	\centering
	\caption[table_total]{{Rendering performance comparison on EuRoC and TUM RGB-D datasets.} 
		\setlength{\fboxsep}{1pt}
		The best three results are highlighted in \colorbox{greenish!40}{\textbf{first}}, \colorbox{lightgreen!40}{second}, and \colorbox{lightyellow}{third}. For algorithms where pose estimation failed, the map could not be reconstructed, and thus the rendering performance value was marked as -.}
	\label{tab:quantitative_render}
	\scriptsize
	\renewcommand{\arraystretch}{1.1}
	\setlength\tabcolsep{3.0pt}
	\begin{tabular} {cc|ccccccccccc|ccc}
		\Xhline{3\arrayrulewidth}
		\multirow{2}{*}{\raisebox{-0.5ex}{\textbf{Method}}} 
		& \multicolumn{1}{c}{\multirow{2}{*}{\raisebox{-0.5ex}{\textbf{Metric}}}}
		& \multicolumn{11}{c}{\textbf{EuRoC}~\cite{burri2016euroc}} 
		& \multicolumn{3}{c}{\textbf{TUM RGB-D~\cite{sturm12iros}}} \\
		\cmidrule(lr){3-14} \cmidrule(lr){14-16}
		& & MH01 & MH02 & MH03 & MH04 & MH05 & V101 & V102 & V103 & V201 & V202 & V203 
		& fr1/desk & fr2/xyz & fr3/office \\
		\Xhline{2\arrayrulewidth}
		\multirow{3}{*} {MonoGS~\cite{matsuki2024gaussian}}
		& PSNR [dB] $\uparrow$ & 17.092 & 17.513 & 10.486 & 18.231 & 13.628 & 19.527 & 14.439 & 11.660 & 18.595 & 10.737 & 11.626 & 21.206 & \cellcolor{lightyellow}22.531 & \cellcolor{lightyellow}22.673 \\
		& SSIM $\uparrow$ & 0.583 & 0.603 & 0.357 & 0.673 & 0.439 & 0.721 & 0.636 & 0.632 & 0.719 & 0.527 & 0.610 & 0.712 & \cellcolor{lightyellow}0.731 & \cellcolor{lightyellow}0.773 \\
		& LPIPS $\downarrow$ & 0.507 & 0.485 & 0.698 & 0.443 & 0.638 & 0.464 & 0.601 & 0.666 & 0.431 & 0.719 & 0.679 & 0.340 & 0.283 & 0.336 \\
		\specialrule{0.5pt}{1pt}{1pt}
		\multirow{3}{*} {MM3DGS-SLAM~\cite{sun2024mm3dgs}}
		& PSNR [dB] $\uparrow$ & 16.918 & 18.367 & - & - & 19.682 & - & 19.880 & - & - & 19.089 & - & 15.015 & 14.482 & 16.920 \\
		& SSIM $\uparrow$ & 0.564 & 0.612 & - & - & 0.702 & - & 0.745 & - & - & 0.752 & - & 0.596  & 0.540 & 0.609 \\
		& LPIPS $\downarrow$ & 0.471 & 0.430 & - & - & 0.386 & - & 0.421 & - & - & 0.400 & - & 0.455  & 0.494 & 0.487 \\
		\specialrule{0.5pt}{1pt}{1pt}
		\multirow{3}{*} {Photo-SLAM~\cite{hhuang2024photoslam}}
		& PSNR [dB] $\uparrow$ & 18.611 & 18.851 & \cellcolor{lightyellow}19.382 & \cellcolor{lightyellow}22.355 & \cellcolor{lightyellow}20.570 & \cellcolor{lightyellow}22.941& \cellcolor{lightyellow}23.846 & 18.504 & - & \cellcolor{lightyellow}22.134 & - & \cellcolor{lightyellow}21.431 & 21.216 & 20.683 \\
		& SSIM $\uparrow$ & 0.614 & 0.632 & 0.668 & \cellcolor{lightyellow}0.807 & \cellcolor{lightyellow}0.744 & \cellcolor{lightyellow}0.817 & \cellcolor{lightyellow}0.840 & 0.784 & - & \cellcolor{lightyellow}0.815 & - & \cellcolor{lightyellow}0.714 &  0.690 & 0.719 \\
		& LPIPS $\downarrow$ & 0.448 & 0.431 & 0.410 & \cellcolor{lightyellow}0.255 & \cellcolor{lightyellow}0.362 & \cellcolor{lightyellow}0.259 & \cellcolor{lightyellow}0.271 &  0.470 & - & \cellcolor{lightyellow}0.261 & - & \cellcolor{lightyellow}0.315 & \cellcolor{lightyellow}0.266 & \cellcolor{lightyellow}0.295 \\
		\specialrule{0.5pt}{1pt}{1pt}
		\multirow{3}{*} {HI-SLAM2~\cite{zhang2025hi}}
		& PSNR [dB] $\uparrow$ & \cellcolor{lightgreen!40}{24.315} & \cellcolor{lightgreen!40}24.410 & \cellcolor{lightgreen!40}22.681 & \cellcolor{lightgreen!40}{23.642} & \cellcolor{lightgreen!40}23.888 & \cellcolor{greenish!40}\textbf{25.139} & \cellcolor{greenish!40}\textbf{26.175} & \cellcolor{greenish!40}\textbf{21.404} & \cellcolor{lightgreen!40}{26.157} & \cellcolor{lightgreen!40}23.513 & \cellcolor{greenish!40}\textbf{24.574} & \cellcolor{lightgreen!40}22.083 & \cellcolor{lightgreen!40}23.976 & \cellcolor{greenish!40}\textbf{24.537} \\
		& SSIM $\uparrow$ & \cellcolor{lightgreen!40}0.826 & \cellcolor{lightgreen!40}0.827 & \cellcolor{lightgreen!40}0.806 & \cellcolor{lightgreen!40}0.843 & \cellcolor{lightgreen!40}0.842 & \cellcolor{lightgreen!40}{0.891} & \cellcolor{greenish!40}\textbf{0.895} & \cellcolor{lightgreen!40}{0.840} & \cellcolor{greenish!40}\textbf{0.888} & \cellcolor{lightgreen!40}0.855 & \cellcolor{greenish!40}\textbf{0.873} & \cellcolor{lightgreen!40}0.750 & \cellcolor{lightgreen!40}0.808 & \cellcolor{greenish!40}\textbf{0.842} \\
		& LPIPS $\downarrow$ & \cellcolor{lightgreen!40}0.190 & \cellcolor{lightgreen!40}0.191 & \cellcolor{lightgreen!40}0.207 & \cellcolor{lightgreen!40}0.203 & \cellcolor{lightgreen!40}0.213 & \cellcolor{lightgreen!40}0.169 & \cellcolor{greenish!40}\textbf{0.173} & \cellcolor{lightgreen!40}0.313 & \cellcolor{lightgreen!40}0.145 & \cellcolor{lightgreen!40}0.206 & \cellcolor{greenish!40}\textbf{0.246} & \cellcolor{lightgreen!40}0.252 & \cellcolor{lightgreen!40}0.174 & \cellcolor{lightgreen!40}{0.177} \\
		\specialrule{0.5pt}{1pt}{1pt}
		\multirow{3}{*} {WildGS-SLAM~\cite{Zheng2025WildGS}}
		& PSNR [dB] $\uparrow$ & \cellcolor{lightyellow}19.573 & \cellcolor{lightyellow}19.629 & 18.803 & 20.167 & 20.541 & 21.941 & 22.691 & \cellcolor{lightyellow}20.867 & \cellcolor{lightyellow}24.137 & 20.874 & \cellcolor{lightyellow}21.126 & 21.169 & 22.462 & 21.752 \\
		& SSIM $\uparrow$ & \cellcolor{lightyellow}0.666 & \cellcolor{lightyellow}0.653 & \cellcolor{lightyellow}0.670 & 0.732 & 0.730 & 0.793 & 0.822 & \cellcolor{lightyellow}0.822 & \cellcolor{lightyellow}0.848 & 0.798 & \cellcolor{lightyellow}0.811 & 0.702 & 0.730 & 0.761 \\
		& LPIPS $\downarrow$ & \cellcolor{lightyellow}0.381 & \cellcolor{lightyellow}0.398 & \cellcolor{lightyellow}0.402 & 0.369 & 0.376 & 0.319 & 0.318 & \cellcolor{lightyellow}0.398 & \cellcolor{lightyellow}0.224 & 0.337 & \cellcolor{lightyellow}0.397 & 0.348 & 0.305 & 0.331 \\
		\specialrule{0.5pt}{1pt}{1pt}
		\multirow{3}{*} {\textbf{Ours}}
		& PSNR [dB] $\uparrow$ & \cellcolor{greenish!40}\textbf{25.220} & \cellcolor{greenish!40}\textbf{25.942} & \cellcolor{greenish!40}\textbf{23.236} & \cellcolor{greenish!40}\textbf{25.121} & \cellcolor{greenish!40}\textbf{25.040} & \cellcolor{lightgreen!40}25.083 & \cellcolor{lightgreen!40}24.965 & \cellcolor{lightgreen!40}21.228 & \cellcolor{greenish!40}\textbf{26.274} & \cellcolor{greenish!40}\textbf{24.408} & \cellcolor{lightgreen!40}23.591 & \cellcolor{greenish!40}\textbf{22.871} & \cellcolor{greenish!40}\textbf{24.422} & \cellcolor{lightgreen!40}23.764 \\
		& SSIM $\uparrow$ & \cellcolor{greenish!40}\textbf{0.865} & \cellcolor{greenish!40}\textbf{0.877} & \cellcolor{greenish!40}\textbf{0.850} & \cellcolor{greenish!40}\textbf{0.875} & \cellcolor{greenish!40}\textbf{0.877} & \cellcolor{greenish!40}\textbf{0.895} & \cellcolor{lightgreen!40}0.873 & \cellcolor{greenish!40}\textbf{0.847} & \cellcolor{lightgreen!40}0.886 & \cellcolor{greenish!40}\textbf{0.879} & \cellcolor{lightgreen!40}0.855 & \cellcolor{greenish!40}\textbf{0.790} & \cellcolor{greenish!40}\textbf{0.836} & \cellcolor{lightgreen!40}0.819 \\
		& LPIPS $\downarrow$ & \cellcolor{greenish!40}\textbf{0.131} & \cellcolor{greenish!40}\textbf{0.130} & \cellcolor{greenish!40}\textbf{0.142} & \cellcolor{greenish!40}\textbf{0.136} & \cellcolor{greenish!40}\textbf{0.135} & \cellcolor{greenish!40}\textbf{0.138} & \cellcolor{lightgreen!40}0.191 & \cellcolor{greenish!40}\textbf{0.284} & \cellcolor{greenish!40}\textbf{0.115} & \cellcolor{greenish!40}\textbf{0.165} & \cellcolor{lightgreen!40}0.256 & \cellcolor{greenish!40}\textbf{0.202} & \cellcolor{greenish!40}\textbf{0.123} & \cellcolor{greenish!40}\textbf{0.163} \\
		\Xhline{3\arrayrulewidth}
	\end{tabular}
	\vspace{-0.4cm}
\end{table*}

\begin{figure*}[ht]
	\centering
	\graphicspath{{pics/exp/}} 
	\includegraphics[width=0.9\textwidth]{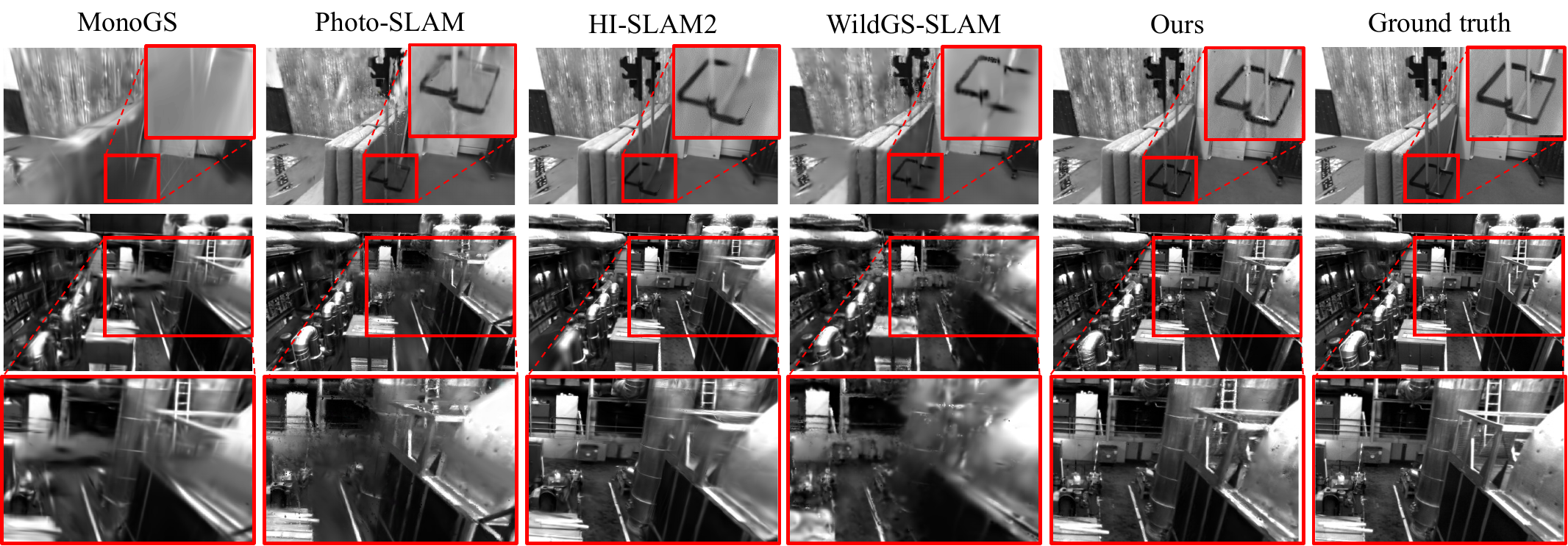}
	\caption{{Qualitative comparison on the EuRoC dataset.} By exploiting optical flow for geometric learning, our approach preserves fine object boundaries and achieves more detailed reconstructions compared with other methods.} 
	\label{fig:result1}
	\vspace{-0.6cm}
\end{figure*}


\vspace{-0.4cm}
\subsection{Qualitative Comparison}
We report qualitative results from reconstructed 3DGS, highlighting the differences between our GaussianFlow SLAM and other algorithms. As shown in~\figref{fig:result2}, compared with two algorithms that utilize the depth prior as a loss term for 3DGS training, our method produces more geometrically valid rasterized depth. 
Furthermore,~\figref{fig:result1} demonstrates that our approach successfully reconstructs detailed structural characteristics of objects. In contrast, Photo-SLAM fails to capture sufficient details due to feature sparsity, while HI-SLAM2 and WildGS-SLAM miss such structures because of the errors coming from the depth prior.
\vspace{-0.4cm}


\subsection{Ablation Study}\label{ablation}

\tabref{abl:pose_module} ablates the effect of GaussianFlow guidance on EuRoC dataset, where large-scale sequences and rapid motions make pose estimation challenging. GaussianFlow guidance improves pose accuracy on most sequences, indicating that leveraging the shared 3DGS geometry provides a benefit for pose estimation.
In addition, \tabref{abl:densify} evaluates the proposed densification and pruning on rendering and pose accuracy, averaged over each dataset. Replacing our modules with the original densification and pruning modules (ODP)~\cite{kerbl20233d} used in all baselines degrades both rendering quality and pose accuracy, mainly due to many small floaters produced during optimization. To avoid excessive Gaussian growth or pose divergence observed with low thresholds in ODP, we set the densification threshold to $0.003$. The performance drop confirms that our error-based densification and pruning are essential. We encourage readers to refer to the supplementary video for qualitative comparisons.

\begin{table}[t]
	\centering
	\caption[Ablation analysis of each module for pose accuracy]{Ablation analysis of GaussianFlow guidance for pose accuracy on each sequence of EuRoC dataset~\cite{burri2016euroc}. (RMSE ATE [m], GFG: GaussianFlow guidance for DBA)}
	\label{abl:pose_module} 
	\resizebox{\columnwidth}{!}{
		\renewcommand{\arraystretch}{1.3}
		\Large
		\begin{tabular}{cccccccccccc}
			\hhline{============}
			& M01 & M02 & M03 & M04 & M05 & V11 & V12      & V13 & V21 & V22 & V23 \\ 
			\Xhline{1.5pt}
			w/o GFG                                                    &  0.041 & {0.036} & 0.047 & 0.143 &  0.060 & 0.088  & 0.068  & 0.073 & 0.066 &  0.060  &  0.069 \\ \cline{1-1}
			\textbf{Ours} & \textbf{0.027} & \textbf{0.035} & \textbf{0.033} & \textbf{0.067} & \textbf{0.053} & \textbf{0.079} & \textbf{0.050} & \textbf{0.051} & \textbf{0.050} & \textbf{0.040} & \textbf{0.066} \\ \cline{1-1}
			\hhline{============}
		\end{tabular}
	}
\vspace{-0.3cm}
\end{table}
\begin{table}[t]
	\centering
	\caption[Ablation analysis of densification and pruning modules.]{Ablation analysis of densification and pruning modules with average quantitative performances. (ODP: original densification and pruning module in 3DGS)}
	\label{abl:densify}
	\resizebox{\columnwidth}{!}{
	\renewcommand{\arraystretch}{1.3}
\begin{tabular}{ccccccccc}
	\hhline{=========}
	& \multicolumn{4}{c|}{EuRoC~\cite{burri2016euroc}}               & \multicolumn{4}{c}{TUM RGB-D~\cite{sturm12iros}} \\ \cline{2-9} 
	& PSNR$\uparrow$ & SSIM$\uparrow$ & {LPIPS$\downarrow$} & \multicolumn{1}{c|}{RMSE$\downarrow$}  & PSNR$\uparrow$   & SSIM$\uparrow$  & LPIPS$\downarrow$ & RMSE$\downarrow$ \\ 			\Xhline{0.9pt}
	ODP & 20.704 & 0.725 & 0.351 & 0.070  &  23.502  & 0.808 & 0.197 & 0.020 \\ \cline{1-1}
	\textbf{Ours} &  \textbf{24.555}  & \textbf{0.871} & \textbf{0.166} &  \textbf{0.050}   & \textbf{23.686} & \textbf{0.815} & \textbf{0.163} & \textbf{0.013} \\
	\hhline{=========}
\end{tabular}
	}
\vspace{-0.7cm}
\end{table}
\vspace{-0.4cm}

\subsection{Runtime and Scalability}\label{runtime_analysis}

\begin{table}[t]
	\centering
	\caption{Runtime comparison on MH05 of EuRoC dataset. Average FPS is computed as the total number of processed images divided by the total wall-clock time. Rendering FPS reports 3DGS rasterization speed.}
	\label{tab:runtime_cmp}
	\setlength{\tabcolsep}{6pt}
	\begin{tabular}{lcc}
		\hline
		Method & Average FPS $\uparrow$ & Rendering FPS $\uparrow$ \\
		\hline
		GaussianFlow SLAM (ours) & 0.17 & 466 \\
		MonoGS~\cite{matsuki2024gaussian} & 3.44 & 235 \\
		MM3DGS-SLAM~\cite{sun2024mm3dgs} & 0.24 & 211 \\
		Photo-SLAM~\cite{hhuang2024photoslam} & 18.59 & 788 \\
		HI-SLAM2~\cite{zhang2025hi} & 4.84 & 770 \\
		WildGS-SLAM~\cite{Zheng2025WildGS} & 0.83 & 461 \\
		\hline
	\end{tabular}
	\vspace{-0.2cm}
\end{table}

\begin{table}[t]
	\centering
	\caption{Runtime breakdown of GaussianFlow SLAM on MH05 of EuRoC dataset. We report average computation time per frame.}
	\label{tab:runtime_breakdown}
	\setlength{\tabcolsep}{6pt}
	\begin{tabular}{lc}
		\hline
		Component module & Time per frame (ms) $\downarrow$ \\
		\hline
		Initial pose optimization & 2,050 \\
		3DGS map optimization & 3,605 \\
		GaussianFlow guidance & 212 \\
		DBA & 58 \\
		\hline
		Total & 5,925 \\
		\hline
	\end{tabular}
	\vspace{-0.6cm}
\end{table}

\begin{table}[t]
	\centering
	\caption{Scalability under map growth on MH05 of EuRoC dataset. 
		We bin the SLAM progress by the total number of Gaussians $N_t$, and report the {cumulative} number of keyframes and the average FPS in each bin.}
	\label{tab:scal_gauss_bins}
	\setlength{\tabcolsep}{6pt}
	\begin{tabular}{lccc}
		\hline
		$N_t$ bin (\# Gaussians) & Cumulative \# keyframes & Average FPS $\uparrow$ \\
		\hline
		$[0,150\text{k})$ & 48 & 0.29 \\
		$[150\text{k},300\text{k})$ & 106 & 0.17 \\
		$[300\text{k},490\text{k})$ & 246 & 0.14 \\
		\hline
	\end{tabular}
	\vspace{-0.7cm}
\end{table}


We report runtime and scalability analysis on the MH05 sequence of the EuRoC dataset under the same evaluation setting. \tabref{tab:runtime_cmp} summarizes the runtime and rendering FPS of GaussianFlow SLAM and prior methods. Although 3DGS rendering itself is not a bottleneck in our method (466 FPS), the online processing speed is 0.17 FPS, lower than other methods. This is largely due to our tightly-coupled alternating design, which repeatedly interleaves 3DGS map optimization with multi-pose optimization. While this alternating loop becomes the main runtime bottleneck, it also enables us to exploit meaningful 3DGS-to-pose feedback, which we found beneficial for pose refinement. \tabref{tab:runtime_breakdown} shows that the dominant cost comes from the 3DGS optimization modules within this loop.
To characterize scalability as the map grows, we partition the sequence into bins based on the total number of Gaussians $N_t$ and report the cumulative number of keyframes for indicating SLAM progress together with the average FPS within each bin, as shown in \tabref{tab:scal_gauss_bins}. We observe that FPS gradually decreases with increasing $N_t$, suggesting that further efficiency gains could be obtained via more selective updates over Gaussians by reducing backward overhead in the kernels. In our setting, the map reaches $\sim$490k Gaussians with an 11\% keyframe ratio.

\vspace{-0.2cm}


\section{Conclusion}\label{sec:concl}
We presented GaussianFlow SLAM, the first monocular 3DGS-SLAM to integrate optical flow supervision with closed-form analytic gradients. Our method not only corrects 3DGS geometry but also demonstrates the benefit of GaussianFlow as a geometric cue in SLAM, achieving state-of-the-art performances.
Nonetheless, the system remains compute-intensive due to repeated first-order 3DGS optimization in the tightly-coupled alternating loop, making it not real-time and necessitating reliance on the DBA module. Future work will explore second-order optimization in 3DGS to improve efficiency and reduce these dependencies.

\vspace{-0.4cm}




\bibliographystyle{URL-IEEEtrans}

\bibliography{URL-bib}

\end{CJK}
\end{document}